\documentclass[lettersize,journal]{IEEEtran}
\usepackage{amsmath,amsfonts}
\usepackage{algorithmic}
\usepackage{algorithm}
\usepackage{array}
\usepackage[caption=false,font=normalsize,labelfont=sf,textfont=sf]{subfig}
\usepackage{textcomp}
\usepackage{stfloats}
\usepackage{subcaption}
\usepackage{url}
\usepackage{verbatim}
\usepackage{graphicx}
\usepackage{cite}
\usepackage[table,dvipsnames]{xcolor}
\usepackage{booktabs}
\usepackage{threeparttable}
\usepackage{siunitx}
\usepackage{adjustbox}
\usepackage{float}
\usepackage{amssymb}  
\usepackage{pifont}
\usepackage{xcolor}
\usepackage{tikz}
\usepackage{hyperref}
\usepackage[utf8]{inputenc}
\DeclareUnicodeCharacter{266B}{$\ musicalnote $}

\newcommand{\warn}{
    \tikz[baseline=0.5ex, scale=0.7]{
        \draw[fill=yellow!80!orange, draw=orange!90!black, thick]
            (0,0) -- (0.22,0.38) -- (0.44,0) -- cycle;
        \node at (0.22,0.14) {\scriptsize\textbf{!}};
    }
}

\newcommand{\correct}{\textcolor{green!60!black}{\checkmark}}
\newcommand{\wrong}{\textcolor{red!70!black}{\text{\sffamily X}}}

\usepackage[most]{tcolorbox}
\include{definitions}

\begin{document}

\title{TDA-RC: Task-Driven Alignment for Knowledge-Based Reasoning Chains in Large Language Models}

\author{Jiaquan Zhang$^{\dagger}$, Qigan Sun$^{\dagger}$, Chaoning Zhang*,~\IEEEmembership{Senior Member, IEEE,} Xudong Wang, Zhenzhen Huang, Yitian Zhou, Pengcheng Zheng, Chi-lok Andy Tai, Sung-Ho Bae,~\IEEEmembership{Member, IEEE,} Zeyu Ma, Caiyan Qin, Jinyu Guo, Yang Yang,~\IEEEmembership{Senior Member, IEEE,} Hengtao Shen,~\IEEEmembership{Fellow, IEEE}

\thanks{Jiaquan Zhang, Zhenzhen Huang, and Jinyu Guo are with School of Information and Software Engineering, University of Electronic Science and Technology of China, Chengdu, China (email: jiaquanzhang2005@gmail.com; alley10086@gmail.com; guojinyu@uestc.edu.cn).
Chaoning Zhang, Yitian Zhou, Pengcheng Zheng, Zeyu Ma, and Yang Yang are with School of Computer Science and Engineering, University of Electronic Science and Technology of China, Chengdu, China (email: chaoningzhang1990@gmail.com; ytzhouuu@gmail.com; zpc777@std.uestc.edu.cn; mazeyu@uestc.edu.cn; yang.yang@uestc.edu.cn).
Chi-lok Andy Tai is with College of Professional and Continuing Education, The Hong Kong Polytechnic University, Hong Kong, China (email: andy.tai@cpce-polyu.edu.hk).
Caiyan Qin is with School of Robotics and Advanced Manufacture, Harbin Institute of Technology, Shenzhen, China (email: qincaiyan@hit.edu.cn). 
Qigan Sun, Xudong Wang, and Sung-Ho Bae are with the School of Computing, 
Kyung Hee University, Yongin-si, South Korea (email: sunqigan@gmail.com; wl200203@khu.ac.kr; shbae@khu.ac.kr). 
Hengtao Shen is with School of Computer Science and Technology, Tongji University, Shanghai, China (e-mail: shenhengtao@hotmail.com).}
\thanks{$^{\dagger}$ These authors contributed equally to this work (co–first authors).}
\thanks{* Corresponding Author}
}


\maketitle

\begin{abstract}
Enhancing the reasoning capability of large language models (LLMs) remains a core challenge in natural language processing. The Chain-of-Thought (CoT) paradigm dominates practical applications for its single-round efficiency, yet its reasoning chains often exhibit logical gaps. While multi-round paradigms like Graph-of-Thoughts (GoT), Tree-of-Thoughts (ToT), and Atom of Thought (AoT) achieve strong performance and reveal effective reasoning structures, their high cost limits practical use. To address this problem, this paper proposes a topology-based method for optimizing reasoning chains. The framework embeds essential topological patterns of effective reasoning into the lightweight CoT paradigm. Using persistent homology, we map CoT, ToT, and GoT into a unified topological space to quantify their structural features. On this basis, we design a unified optimization system: a Topological Optimization Agent diagnoses deviations in CoT chains from desirable topological characteristics and simultaneously generates targeted strategies to repair these structural deficiencies. Compared with multi-round reasoning methods like ToT and GoT, experiments on multiple datasets show that our approach offers a superior balance between reasoning accuracy and efficiency, showcasing a practical solution to ``single-round generation with multi-round intelligence''.
\end{abstract}


\section{Introduction}
With the widespread application of LLMs~\cite{ma2025general,huang2022towards} to complex tasks such as mathematical and logical reasoning~\cite{xu2025large}, evaluating and optimizing the quality of their generative reasoning processes has become a research focus. Among various reasoning paradigms, CoT~\cite{wei2022chain} and its variants significantly enhance reasoning ability by introducing intermediate steps~\cite{li2025exploring,zhong2025large,qiu2025explainable,nguyen2025improving}. However, due to its single-round generation and ease of deployment, CoT has become the dominant paradigm in industrial and practical applications. Unfortunately, its generated reasoning chains often suffer from inconsistency, lack of self-verification, and insufficient reasoning depth.

Meanwhile, the research community has introduced more sophisticated reasoning paradigms such as ToT~\cite{yao2023tree}, GoT~\cite{besta2024graph}, and AoT~\cite{teng2025atom}. Through iterative exploration, branching, and cyclic verification, these frameworks achieve remarkable performance across multiple benchmarks. However, their deployment complexity, high computational cost, and long response latency hinder practical adoption~\cite{stechly2024chain,panfasttree,jin2024large}. This creates a core dilemma: although more effective reasoning paradigms exist, practical constraints force continued reliance on the simpler yet structurally flawed CoT~\cite{xu2025large}. This raises a fundamental question: Can we make a better trade-off between reasoning performance and resources? Paradigms like ToT, GoT, and AoT suggest that high-quality reasoning is not merely a linguistic process but also a structural phenomenon, whose effectiveness arises from its internal connectivity and organization. Building on this insight, we consider whether the topological characteristics of reasoning processes can reveal the underlying principles of successful inference. In other words, \textbf{Can we extract the topological features that drive successful reasoning from complex multi-round processes and transfer them into a practical CoT framework, without incurring costly multi-turn interactions?} 

To achieve this goal, we adopt Topological Data Analysis (TDA) as a tool for the structured analysis of thought chains. TDA~\cite{chazal2021introduction,uchendu2024unveiling} provides a powerful mathematical framework for understanding complex structural patterns. In particular, persistent homology captures features such as loops, branches, and connectivity components, which makes it naturally suited for quantifying the structural properties of reasoning chains. We employ persistent homology to map different reasoning structures into a unified topological space, where their structural characteristics, such as branching, cyclicity, and connectivity, are quantitatively represented. Based on this topological representation, we design a unified optimization system: a Topological Optimization Agent. This agent leverages prior knowledge and the topological representations derived from TDA to generate a structured diagnostic report, assessing how a CoT reasoning chain deviates from high-quality topological patterns. The agent then formulates targeted strategies to repair structural deficiencies based on this report and produces executable optimization prompts. These prompts are guided by a mapping from deviation patterns to optimization templates, enabling adaptive correction of reasoning structures. The main contributions are as follows:

\begin{itemize}
    \item  We propose a topology-based unified representation method for reasoning chains grounded in persistent homology, which extracts structural features underlying successful reasoning from complex multi-round processes and leverages them to enhance chain-of-thought reasoning.
    \item  We design an optimization framework that integrates topological priors with LLMs' semantic generation in a closed-loop process, from diagnosis to repair. The agent adaptively refines evaluation criteria and optimization strategies for continuous improvement.
    \item  We extensively evaluate TDA-RC and find that leveraging topological features for reasoning optimization substantially improves the accuracy–efficiency trade-off, with normalized structural metrics and ablation studies validating its effectiveness and interpretability.
\end{itemize}
Following this ``analyze complex structures, optimize simple structures'' paradigm, this study aims to push the intrinsic quality limits of CoT and advance prompt engineering from heuristic craft toward a data-driven structural science.

\section{Related Work}
\subsection{Chain-of-Thought's Origin and Evaluation Methods.}
Chain-of-Thought (CoT) reasoning, which decomposes complex problems into intermediate reasoning steps, has been widely applied across various reasoning tasks~\cite{chang2024survey, chen2025towards, yu2023towards,li2025exploring}. First introduced by Wei et al.~\cite{wei2022chain}, CoT significantly improved large language models’ (LLMs) performance on arithmetic, commonsense, and symbolic reasoning by prompting step-by-step inference. Building on this idea, several structured reasoning paradigms have emerged. Yao et al.~\cite{yao2023tree} proposed Tree-of-Thought (ToT), which explores multiple reasoning paths in a tree structure and allows self-assessment of intermediate progress. Besta et al.~\cite{besta2024graph} extended this concept to Graph-of-Thought (GoT), representing reasoning as a graph with flexible connections and cycles. Li et al.~\cite{li2025syzygy} further enhanced LLM reasoning accuracy by introducing auxiliary reasoning paths. With the widespread adoption of CoT-based reasoning, evaluating the quality of reasoning chains has become an open challenge. Most existing methods assess only final-answer accuracy while neglecting structural properties such as connectivity, stability, and logical consistency. Turpin et al.~\cite{turpin2023language} noted that CoT explanations often exhibit unfaithfulness, and Valmeekam et al.~\cite{sprague2023musr} criticized static evaluations that overlook reasoning stability. Similarly, Wu et al.~\cite{wu2024mitigating} and Chen et al.~\cite{chen2025towards} emphasized that ignoring path depth and intermediate consistency can underestimate reasoning errors. To address these gaps, recent works aim to evaluate reasoning processes more directly. Liu et al.~\cite{nguyen2024direct} proposed a knowledge-graph-based framework to analyze reasoning fidelity. Zhou et al.~\cite{zhou2025dissecting} provided a multidimensional evaluation of logical reasoning abilities, while Jacovi et al.~\cite{jacovi2024chain} introduced the REVEAL benchmark for assessing reasoning validity and consistency. Although these approaches improve reasoning evaluation, they mainly focus on local step analysis and lack a global perspective on the structural complexity of reasoning chains. These limitations highlight the need for a new framework that goes beyond final-answer accuracy and captures the global structural properties of reasoning chains, including their complexity, stability, and coherence.

\subsection{Topological Data Analysis Across Fields.}
Topological Data Analysis (TDA) is a topology-based technique designed to extract and analyze the geometric and structural organization of data~\cite{chazal2021introduction,malott2022survey}. Unlike conventional data analysis methods that focus on individual attributes or samples, TDA emphasizes both global and local structures, capturing the relationships and connectivity patterns within complex systems. Its core tool, Persistent Homology~\cite{weinberger2011persistent,germain2024persistence,guo2018sparse}, identifies multi-scale topological features such as connectivity, loops, and voids, which are represented as persistence diagrams or barcodes~\cite{edelsbrunner2012persistent}. These representations provide a compact and noise-robust summary of data topology, enabling the study of structural invariances and variations~\cite{edelsbrunner2008persistent, okediji2024persistent}. TDA has been successfully applied across diverse domains. In biology, it has aided gene expression analysis, protein folding prediction, and clinical outcome modeling~\cite{amezquita2020shape, iniesta2022topological}. In physics and materials science, it has revealed topological patterns in turbulence and microstructure dynamics~\cite{nauleau2022topological, yang2024topological, wang2025persistent}. In computer science, TDA supports data mining, machine learning, and image analysis tasks~\cite{tu2019efficient, adjei2023topological}, offering a structure-aware perspective for feature extraction and dimensionality reduction~\cite{leykam2023topological}. Building on these advances, we extend TDA to the domain of reasoning-chain analysis. By applying persistent homology to reasoning structures, we model their hierarchical organization, quantify connectivity and stability, and detect potential weaknesses or error propagation within the reasoning process. This provides a principled, topology-driven foundation for evaluating and optimizing reasoning quality.

\section{Method}
\begin{figure*}[ht!]
    \centering
    \includegraphics[width=1\linewidth]{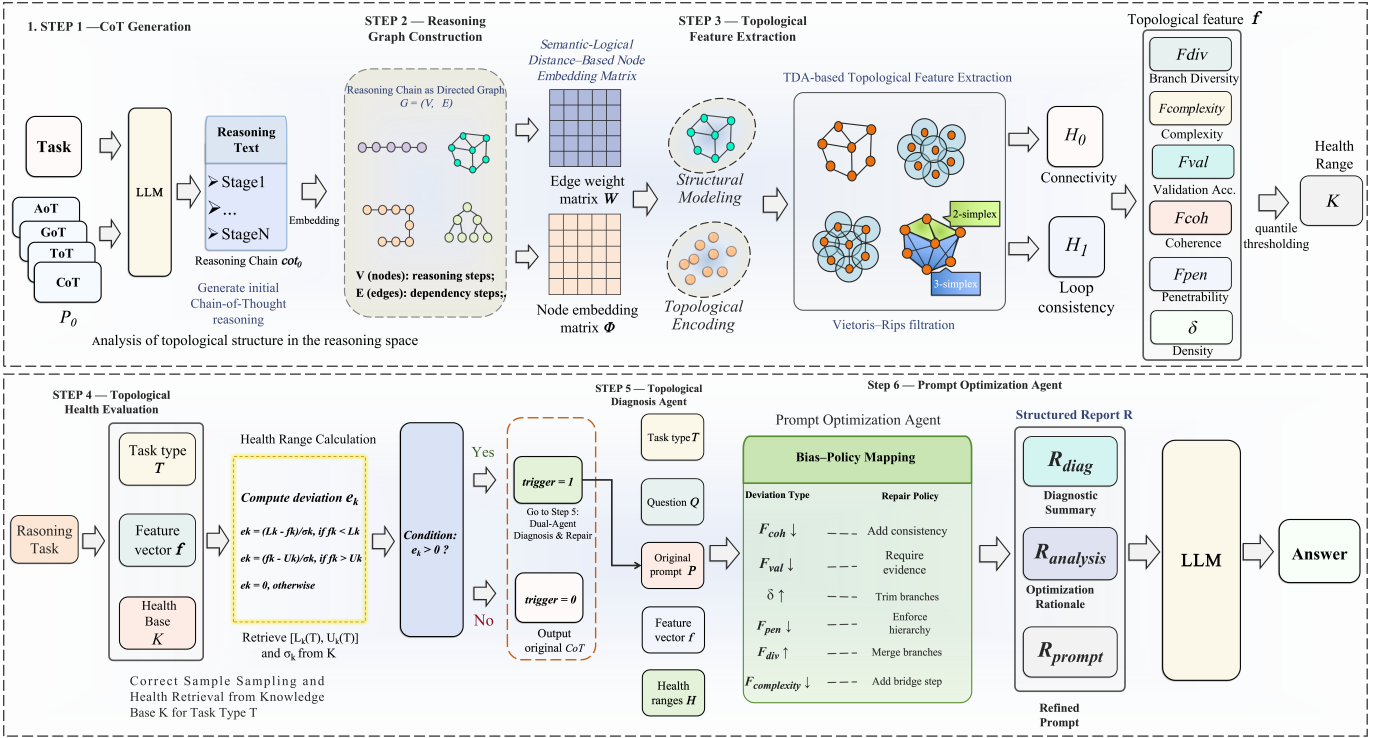}
    \caption{The upper part illustrates the offline procedure for constructing task-specific health bands, where reasoning chains are modeled as topological graphs and persistent homology features are extracted to characterize their structural properties. The lower part presents the online usage pipeline of TDA-RC, which performs a single-round closed-loop optimization consisting of health evaluation, structural diagnosis, and agent-based prompt repair. By combining persistent homology with reasoning feature analysis, TDA-RC identifies structural inconsistencies in LLM reasoning and generates targeted repair strategies, enabling one-shot refinement of the prompt to improve structural coherence and task accuracy.}
    \label{fig:tda_rc_overview}
\end{figure*}

We propose TDA-RC, a topology-aware, agent-collaborative framework to optimize LLM reasoning chains. TDA-RC first diagnoses the model’s initial CoT; if key topological indicators fall below task-specific bands, it injects structure-aware refinements guided by TDA signals and regenerates the reasoning once. An overview of the pipeline is shown in Figure~\ref{fig:tda_rc_overview}. This section is organized into five parts introducing the theoretical foundations and implementation details of TDA-RC: 
(i) Topological Data Analysis Foundations in Section~\ref{sec:tda_Foundations}; 
(ii) Graph Construction in Section~\ref{sec:tda}; 
(iii) Topological Indicators in Section~\ref{sec:tda_Topological_Indicators}; 
(iv) Topological Health Modeling in Section~\ref{sec:health}; 
and (v) Agent-based Optimization and Repair in Section~\ref{sec:agents}.

\subsection{Topological Data Analysis Foundations}
\label{sec:tda_Foundations}
This module is dedicated to explaining the theoretical foundations of our topological diagnostic framework. It primarily consists of two key components: the concept of persistent homology and the Vietoris--Rips filtration, which allows us to capture multi-scale structural properties of reasoning chains, and the semantic--logical composite manifold, which combines semantic and logical features to provide a unified space for topological analysis.

\subsubsection{Persistent Homology and Vietoris--Rips Complex}
TDA provides a way to view data not merely as points in space,
but as shapes that carry global structural meaning.
Its core tool, \textit{persistent homology}, describes how the topology of a dataset evolves across multiple scales and reveals which structural patterns remain stable under perturbations. Given a metric space $(X,d)$, we construct a sequence of simplicial complexes parameterized by a scale
$\epsilon$, known as the Vietoris--Rips filtration:
\begin{equation}
\mathcal{VR}_\epsilon(X)=\{\sigma\subseteq X\mid \forall x,y\in\sigma,\, d(x,y)\le\epsilon\}.    
\end{equation}
When $\epsilon$ increases, isolated points gradually merge into clusters, edges and loops appear,
and finally the structure becomes fully connected:
\begin{equation}
\mathcal{VR}_{\epsilon_0}(X)\hookrightarrow
\mathcal{VR}_{\epsilon_1}(X)\hookrightarrow
\cdots\hookrightarrow
\mathcal{VR}_{\epsilon_n}(X).
\end{equation}
Applying homology to the filtration yields a sequence of homology groups and the induced maps between them:
\begin{equation}
H_k(\mathcal{VR}_{\epsilon_0})\to H_k(\mathcal{VR}_{\epsilon_1})\to\cdots\to H_k(\mathcal{VR}_{\epsilon_n}),
\end{equation}
where $H_0$ captures connected components and $H_1$ represents one-dimensional cycles.
The persistence of these features, that is, how long they survive as $\epsilon$ grows,
quantifies their structural stability. In our setting, a reasoning chain can be viewed as
a geometric object embedded in a semantic space: $H_0$ features reflect the number of coherent
reasoning branches, whereas $H_1$ features measure the emergence of reasoning loops, which are
often associated with self-consistency or verification. Long-lived topological features
correspond to stable logical structures, while short-lived ones tend to indicate fragmented
or noisy reasoning paths.

\subsubsection{Semantic--Logical Composite Manifold}
However, reasoning chains are not purely geometric objects; their structure arises from both linguistic semantics and logical relations. To capture this dual nature, we follow the notion of a semantic--logical composite manifold~\cite{uchendu2024unveiling}, which combines two complementary spaces:
\begin{equation}
\begin{aligned}
\mathcal{M}_{\text{semantic}}
    &= \{\phi_{\text{LM}}(v_i) \mid v_i \in V\},\\
\mathcal{M}_{\text{logical}}
    &= \{(d_i, b_i, c_i) \mid v_i \in V\}.
\end{aligned}
\end{equation}
Here, $\phi_{\text{LM}}$ denotes the semantic embedding extracted from a pretrained BERT model. 
The tuple $(d_i, b_i, c_i)$ further encodes the logical attributes of each reasoning step, where 
$d_i$ is its depth in the reasoning hierarchy, $b_i$ is the branch identifier, and $c_i$ measures 
its local structural connectivity. We next formalize these logical components, along with the 
overall distance metric used in our analysis.

\paragraph{Local Connectivity $c_i$}
To quantify the local connectivity of $v_i$, we use a combination of one-hop neighborhood features:
\begin{equation}
c_i = \frac{1}{\beta}
\left(
\alpha_{\text{deg}} \cdot \frac{\deg(v_i)}{\deg_{\max}}
+
\alpha_{\text{cc}} \cdot \text{CC}(v_i)
\right),
\end{equation}
where $\deg(v_i)$ is the degree of $v_i$ (in-degree + out-degree), $\deg_{\max}$ is the maximum 
degree in $G$, and $\text{CC}(v_i)$ is the clustering coefficient among the neighbors of $v_i$. 
The coefficients $\alpha_{\text{deg}}, \alpha_{\text{cc}}, \beta$ control weighting and 
normalization. This formulation follows standard graph-topology practice.

\paragraph{Logical Distance $d_{\text{logical}}(v_i, v_j)$}
The logical distance captures structural dissimilarity between two reasoning steps:
\begin{equation}
\begin{aligned}
d_{\text{logical}}(v_i, v_j)
    &= \gamma_d \, |d_i - d_j|
     + \gamma_b \, \mathbf{1}[b_i \neq b_j] \\
    &\quad + \gamma_c \, |\mathrm{norm}(c_i) - \mathrm{norm}(c_j)|,
\end{aligned}
\end{equation}
where the three terms respectively quantify hierarchical disparity, cross-branch mismatch, 
and differences in local structural roles. The weights satisfy 
$\gamma_d + \gamma_b + \gamma_c = 1$.

\paragraph{Composite Semantic--Logical Distance}
Finally, semantic and logical similarity are combined into a unified distance:
\begin{equation}
\begin{aligned}
d_{\text{composite}}(v_i, v_j)
    &= \lambda_s \, d_{\text{semantic}}(v_i, v_j)
     + \lambda_l \, d_{\text{logical}}(v_i, v_j),\\
\lambda_s + \lambda_l &= 1.
\end{aligned}
\end{equation}
This composite metric ensures that reasoning steps are close either when they convey similar 
content or when they occupy analogous structural roles. Applying Vietoris--Rips filtration to 
this manifold yields persistent barcodes that capture the “shape of reasoning’’ across scales, 
forming the foundation of our topology-aware optimization in Section~\ref{sec:tda}.

\subsection{Graph Construction}
\label{sec:tda}
We represent each reasoning chain as a directed graph \(G=(V,E)\), where each node \(v_i\) denotes a reasoning step
and each edge \(e_{ij}\) encodes a sequential, referential, or dependency relation.
To characterize multi-scale structural properties, we compute persistent homology over a scale sequence \(\{\epsilon_s\}_{s=1}^{S}\), where \(\epsilon_s\) denotes the filtration scale at level \(s\). Here, \(S\) is the total number of filtration steps and is independent of the task type \(T\) used elsewhere; we index levels by \(s\in\{1,\dots,S\}\) with a non-decreasing schedule \(0<\epsilon_1\le\epsilon_2\le\cdots\le\epsilon_S\). We then extract \(H_0\) (connectivity) and \(H_1\) (loop verification) from the Vietoris--Rips complex. These computations rely on a unified node embedding:
\begin{equation}
\Phi(v_i) = \mathrm{MLP}\left([s_i; p_d(d_i); p_b(b_i); p_c(c_i)]\right) \in \mathbb{R}^{256}
\end{equation}
Here, $s_i$ denotes the semantic embedding, and $p_d$, $p_b$, and $p_c$ encode the hierarchical depth, branch identity, and local connectivity statistics. To clarify the role of the MLP in Eq.~(8), we note that the semantic vector $s_i$ (high-dimensional) and the logical indicators $p_d, p_b, p_c$ (low-dimensional and discrete) lie in heterogeneous feature spaces. Direct concatenation would introduce scale imbalance and feature dominance. Thus, the MLP above functions as a structure-preserving projection head rather than a semantic learner. Its weights are orthogonally initialized and remain frozen during the entire process. This design follows the theory of random projections, ensuring that the projected space $\mathbb{R}^{256}$ approximately preserves pairwise distances and the underlying topological structure with high probability (Johnson--Lindenstrauss lemma). The resulting unified node representation supports stable graph construction. Based on this representation, we define the pairwise node metric used to construct the edge weights of the reasoning graph. The edge weighting is defined as follows:
\begin{equation}
w_{ij}=\alpha\, \cos(s_i,s_j) + (1-\alpha)\,\mathrm{1}\,[\text{explicit dependency}], \alpha=0.7.
\end{equation}

To ensure the comparability of topological indicators across different reasoning paradigms, we adopt a unified modeling scheme for explicit dependencies. For multi-round structured paradigms such as ToT, GoT, and AoT, the reasoning process naturally exhibits nonlinear control flows—including tree expansion, graph-level verification, and atom-level recomposition—which inherently provide clear parent–child relations and cross-step validation links. We directly treat these native structural edges as explicit dependencies and incorporate them into Eq. (9) without additional inference, thereby ensuring the authenticity and reproducibility of the structural information.

In contrast to multi-round paradigms such as ToT and GoT, a standard CoT trace is a purely linear sequence of natural-language steps, which collapses into a simple path and yields a topologically degenerate graph. To enable meaningful persistent homology while preserving the semantics of CoT, we instruct the model to explicitly annotate step-level dependencies during CoT generation. Each step outputs its textual content along with the earlier steps it relies on, and we add a directed edge from each referenced step to the current one. This introduces minimal but essential non-sequential links, providing connectivity patterns such as local aggregation and short verification loops that prevent topological collapse. The resulting dependency-enhanced graph preserves the original reasoning flow while enabling CoT and multi-round traces to be embedded in a unified metric space for consistent topological analysis.

\subsection{Topological Indicators}
\label{sec:tda_Topological_Indicators}
Based on node hierarchy, graph structure, and homological features, we define six complementary
topological indicators that jointly characterize the structural health of a reasoning chain. Each
metric reflects a distinct form of reasoning behavior relevant to multi-step reliability.

\paragraph{Validation Rate $F_{\text{val}}$}
The validation rate measures the proportion of reasoning steps that include explicit verification
or evidence-based justification:
\begin{equation}
F_{\text{val}} = \frac{|V_{\text{val}}|}{|V|}.
\end{equation}
Higher values correspond to stronger long-range referencing and evidence aggregation.

\paragraph{Connectivity Density $\delta$}
Connectivity density evaluates the compactness of the reasoning graph and the directness of the
information flow:
\begin{equation}
\delta = \frac{|E|}{|V|(|V|-1)/2}.
\end{equation}
Extremely high or low density indicates either redundant branching or overly collapsed pathways.

\paragraph{Loop Coherence $F_{\text{coh}}$}
Loop coherence assesses the stability and self-consistency of cyclic reasoning patterns, derived
from the persistence of one-dimensional homological features:
\begin{equation}
F_{\text{coh}}
    = \frac{\sum_{j \in H_1} \text{persist}(\beta_j)}
           {\sum_{j \in H_1} \left( \text{birth}(\beta_j) + \text{death}(\beta_j) \right)}.
\end{equation}
Stable short cycles often reflect local verification loops essential in symbolic and multi-hop tasks.

\paragraph{Penetrability $F_{\text{pen}}$}
Penetrability quantifies the hierarchical depth of the reasoning process:
\begin{equation}
F_{\text{pen}} = \frac{1}{|V|} \sum_{i \in V} d_i.
\end{equation}
Tasks requiring multi-hop derivation or arithmetic depth typically exhibit higher penetrability.

\paragraph{Branch Diversity $F_{\text{div}}$}
Branch diversity measures both the number of independent reasoning branches and the balance of the
distribution across them:
\begin{equation}
F_{\text{div}} = \mathrm{Norm}\left( |B| \cdot \mathrm{Entropy}(P_B) \right),
\end{equation}
\begin{equation}
\mathrm{Entropy}(P_B) = -\sum_{b \in B} p_b \log p_b .
\end{equation}
Moderate branching supports exploration, whereas excessive divergence signals unfocused reasoning.

\paragraph{Structural Complexity $F_{\text{complexity}}$}
Structural complexity captures the scale, local connectivity, and global reachability of the
reasoning graph:
\begin{equation}
F_{\text{complexity}}
    = |V| \cdot \mathrm{avg\_deg}(G)
      \cdot \frac{1}{\mathrm{avg\_shortest\_path\_length}}.
\end{equation}
Higher complexity reflects richer, non-trivial reasoning structures beyond shallow linear chains.

\subsection{Topological Health Modeling}
\label{sec:health}

This section explains how to model the health of the reasoning process structure based on topological features and trigger necessary repairs according to task requirements. We extract topological feature statistics from typical tasks across multiple task paradigms and aggregate these features to construct task-specific health standards. Each task type \(T\) induces a set of topological health bands, which describe the structural health standards for that task. Specifically, for each topological metric \(F_k\), we estimate its empirical distribution \(p_k^{(T)}(f)\) and define the task-specific health bands as:
\begin{equation}
H_k^{(T)} = [L_k^{(T)}, U_k^{(T)}],
\end{equation}
where \(L_k^{(T)}\) and \(U_k^{(T)}\) are the lower and upper health thresholds for feature \(F_k\). This band represents the structurally healthy range of feature values for the task. For approximately normal distributions, \(L_k^{(T)}\) and \(U_k^{(T)}\) correspond to \(\mu_k^{(T)} \pm \sigma_k^{(T)}\), where \(\mu_k^{(T)}\) and \(\sigma_k^{(T)}\) represent the mean and standard deviation of the empirical distribution \(p_k^{(T)}(f)\). For skewed or heavy-tailed distributions, we use the interquartile range (IQR) to handle them, where \(L_k^{(T)} = Q_1^{(T)}\) and \(U_k^{(T)} = Q_3^{(T)}\), and \(Q_1^{(T)}\) and \(Q_3^{(T)}\) are the first and third quartiles of \(p_k^{(T)}(f)\). For a given reasoning instance's topological feature vector \(f\), we measure the deviation of each metric:
\begin{equation}
e_k = 
\begin{cases}
\frac{L_k^{(T)} - f_k}{\sigma_k^{(T)}}, & \text{if } f_k < L_k^{(T)}, \\
\frac{f_k - U_k^{(T)}}{\sigma_k^{(T)}}, & \text{if } f_k > U_k^{(T)}, \\
0, & \text{otherwise}.
\end{cases}
\end{equation}
This results in a deviation vector \(e = (e_1, \dots, e_m)\). When any high-priority metric in the task-specific set \(\mathcal{P}_T\) violates its health range, structural repair is triggered:
\begin{equation}
\exists\, k \in \mathcal{P}_T : e_k > 0 \quad \Rightarrow \quad \text{trigger structural repair}.
\end{equation}
We also define the maximum deviation severity \(\|\mathbf{e}\|_{\infty}^{(\mathcal{P}_T)}\), which represents the maximum normalized deviation, limited to the metrics within the task priority set \(\mathcal{P}_T\), as well as the violation count:
\begin{equation}
V = \sum_k \mathbf{1}[\, e_k > 0 \,].
\end{equation}
Through this health modeling, we can identify structural deviations in the reasoning chain and quantify the extent of deviation of each task's topological features from their health bands. Next, we design an agent-based optimization mechanism that leverages topological analysis results to optimize prompts.

\subsection{Agent-based Optimization and Repair}
\label{sec:agents}

\begin{tcolorbox}[
    float,
    floatplacement=h,
    enhanced,
    breakable,
    colback=blue!1,
    colframe=blue!40!black,
    colbacktitle=blue!10!white,
    coltitle=black,
    title={\textbf{Prompt Optimization Agent Prompt Template}},
    fonttitle=\bfseries\small,
    boxrule=0.6pt,
    arc=2pt,
    left=8pt,
    right=8pt,
    top=6pt,
    bottom=6pt,
    borderline west={2pt}{0pt}{blue!60},
    drop shadow southeast,
    sharp corners
]

\ttfamily
\small
\textbf{System Role:} You are the Prompt Optimization Agent. 
Your task is to refine and restructure prompts based on topological reasoning diagnostics.\\[4pt]

\textbf{Input:}\\
\hspace*{1.5em}Task type: \{T\}\\
\hspace*{1.5em}Question: \{Q\}\\
\hspace*{1.5em}Original prompt: \{P\}\\
\hspace*{1.5em}Feature vector: \{$\mathbf{f}$\} (topological reasoning features)\\
\hspace*{1.5em}Health bands: \{$H_k^{(T)}$\} \\[6pt]


\textbf{Output (Structured Report R):}\\
\hspace*{1.5em}$R_{\text{diag}}$: structural diagnosis of the prompt.\\
\hspace*{1.5em}$R_{\text{analysis}}$: rationale for optimization.\\
\hspace*{1.5em}$R_{\text{prompt}}$: optimized prompt.
\end{tcolorbox}
After evaluating the topological health of a reasoning chain and detecting any structural
deviations, we design a prompt-optimization agent that interprets the topological diagnostics and
translates them into targeted prompt refinements. The agent receives the feature vector
\(\mathbf{f}\) and the task-specific health bands \(H_k^{(T)}\), compares each metric against its
acceptable range, and identifies the corresponding structural issues—for example, low coherence
\(F_{\mathrm{coh}}\) indicating missing validation, or high density \(\delta\) suggesting redundant
reasoning paths. Based on these deviations, the agent selects appropriate repair strategies from a small set of topology-informed guidelines. It may insert validation steps to increase coherence, request explicit evidence to improve \(F_{\mathrm{val}}\), prune redundant branches to reduce density, or add intermediate steps to enhance structural complexity. Importantly, while deviation detection is rule-based, the repair stage is generative, that is, the agent functions as an LLM role that generates semantic-preserving refinements to strengthen the reasoning structure.

The final output is a structured report containing (i) a concise diagnosis of structural
weaknesses, (ii) the rationale behind the repairs, and (iii) the optimized prompt used for
regeneration. This ensures that TDA-RC performs a single-pass, topology-guided refinement that
strengthens reasoning coherence without introducing additional multi-round overhead.

\section{Experiments}
\subsection{Experimental Setup}
\subsubsection{Reasoning Baselines} 
To evaluate TDA-RC comprehensively, we compare it against a broad set of existing reasoning paradigms. 
For consistency with our main experimental comparison, we organize the baselines into two categories: 
chain-based reasoning methods and prompt-based optimization methods. 
All method descriptions are provided below.
\paragraph{Chain-based Reasoning Methods}
These approaches primarily operate at the level of reasoning traces, 
by sampling, refining, or structurally exploring intermediate chains.
\begin{itemize}
    \item \textbf{Chain-of-Thought (CoT)}~\cite{wei2022chain}: 
    Generates intermediate reasoning steps to enhance logical consistency in problem-solving.

    \item \textbf{CoT-SC (Self-Consistency, $n=5$)}~\cite{wei2022chain}: 
    Extends CoT by sampling multiple reasoning paths and selecting the most consistent answer among them.

    \item \textbf{Self-Refine}~\cite{madaan2023self}: 
    Enables the model to iteratively review and refine its own initial outputs for improved reasoning accuracy.

    \item \textbf{AFlow}~\cite{zhang2024aflow}: 
    Optimizes the reasoning workflow through Monte Carlo Tree Search (MCTS), 
    providing a more systematic and adaptive reasoning process.

    \item \textbf{Tree-of-Thought (ToT)}~\cite{yao2023tree}: 
    Explores reasoning through a tree-structured process, generating and evaluating multiple branches to improve diversity and robustness.

    \item \textbf{Graph-of-Thought (GoT)}~\cite{besta2024graph}: 
    Represents reasoning as a graph structure that explicitly models inter-step dependencies and cyclic verification.

    \item \textbf{Forest-of-Thought (FoT, $n=8$)}~\cite{bi2024forest}: 
    Enhances reasoning by constructing multiple reasoning trees with sparse activation and dynamic self-correction.
    In our implementation, we set the branch number to 3 to balance diversity and computational cost.

    \item \textbf{Atom-of-Thought (AoT)}~\cite{teng2025atom}: 
    Decomposes tasks into minimal reasoning atoms and recombines them into structured reasoning processes for solving complex problems.
\end{itemize}

\paragraph{Prompt-based Optimization Methods}
These approaches mainly modify the input prompt format, role, or instructional style 
to steer the model toward better reasoning, without introducing an explicit external 
search over many alternative traces.
\begin{itemize}
    \item \textbf{Highlighted Chain-of-Thought (HoT)}~\cite{nguyen2025hot}: 
    Improves reasoning focus by highlighting key facts and emphasizing core elements in each reasoning step.

    \item \textbf{Instruction Induction}~\cite{honovich2022instruction}: 
    Encourages models to induce generalizable reasoning rules from diverse instructional examples.

    \item \textbf{Role/Persona Prompting}~\cite{chen2024persona}: 
    Introduces specific roles or personas to enhance contextual reasoning,
    particularly effective in role-playing and multi-task environments.

    \item \textbf{Prompt Canvas}~\cite{hewing2024prompt}: 
    Organizes multi-step reasoning into a visual or modular prompt canvas to manage and refine reasoning flow.

    \item \textbf{Analogical Prompting}~\cite{wei2022chain}: 
    Leverages analogy examples from similar problems to infer solutions for new tasks.

    \item \textbf{TDA-RC (ours)}: 
    Our proposed topology-guided reasoning optimization framework that diagnoses and repairs structural deficiencies 
    using topological feature deviations.
\end{itemize}

\subsubsection{Datasets} 
We evaluate the reasoning capabilities of models on a diverse set of benchmarks covering mathematical, logical, commonsense, and multi-hop reasoning tasks:
\begin{itemize}
    \item \textbf{MATH}~\cite{hendrycks2021measuring}: 
    A large-scale, high-difficulty mathematical problem set designed to assess symbolic and arithmetic reasoning.

    \item \textbf{OlympiadBench}~\cite{he2024olympiadbench}: 
    A collection of competition–level math problems used to test models’ logical and deductive reasoning skills.

    \item \textbf{GSM8K}~\cite{cobbe2021training}: 
    An elementary arithmetic reasoning dataset that evaluates a model’s step-by-step problem-solving capability.

    \item \textbf{BBH}~\cite{suzgun2022challenging}: 
    A diverse suite of challenging tasks across multiple domains, used to measure general reasoning robustness and transferability.

    \item \textbf{MMLU-CF}~\cite{zhao2024mmlu}: 
    A factual and commonsense reasoning subset of MMLU, designed to evaluate knowledge consistency and logical factual reasoning.

    \item \textbf{LongBench}~\cite{bai2023longbench}: 
    A long-context reasoning benchmark for assessing models’ memory, coherence, and information retention over extended text spans.

    \item \textbf{HotpotQA}~\cite{yang2018hotpotqa}: 
    A multi-hop question-answering dataset that tests the model’s ability to perform cross-sentence reasoning and evidence integration.

    \item \textbf{MuSiQue}~\cite{trivedi2022musique}: 
    A complex question-answering benchmark emphasizing hierarchical reasoning and multi-document evidence aggregation.
\end{itemize}

\subsubsection{Implementation Details.} We conduct the evaluation on three API-based large language models: GPT-4o-mini, Qwen-Turbo, and DeepSeek-V3. GPT-4o-mini~\cite{hurst2024gpt} is an instruction-tuned model with efficient general reasoning capabilities; Qwen-Turbo~\cite{hui2024qwen2} is a high-throughput language model based on a mixture-of-experts architecture; and DeepSeek-V3~\cite{liu2024deepseek} is a multi-stage reasoning model optimized for long-context and logical
tasks. We standardize the inference parameters across all models unless otherwise specified: temperature $= 0.2$, top-$p = 1.0$, frequency\_penalty $= 0$, and presence\_penalty $= 0$. The maximum generation length is determined by the dataset: 512 tokens for MATH, OlympiadBench, GSM8K, BBH, MMLU-CF, HotpotQA, and MuSiQue, and 1024 tokens for LongBench to accommodate long-context reasoning.

\subsection{Analysis of Topological Indicators}
\subsubsection{Structural Convergence and Health-Band Motivation}
\begin{figure}[!h]
    \centering
    \includegraphics[width=1\linewidth]{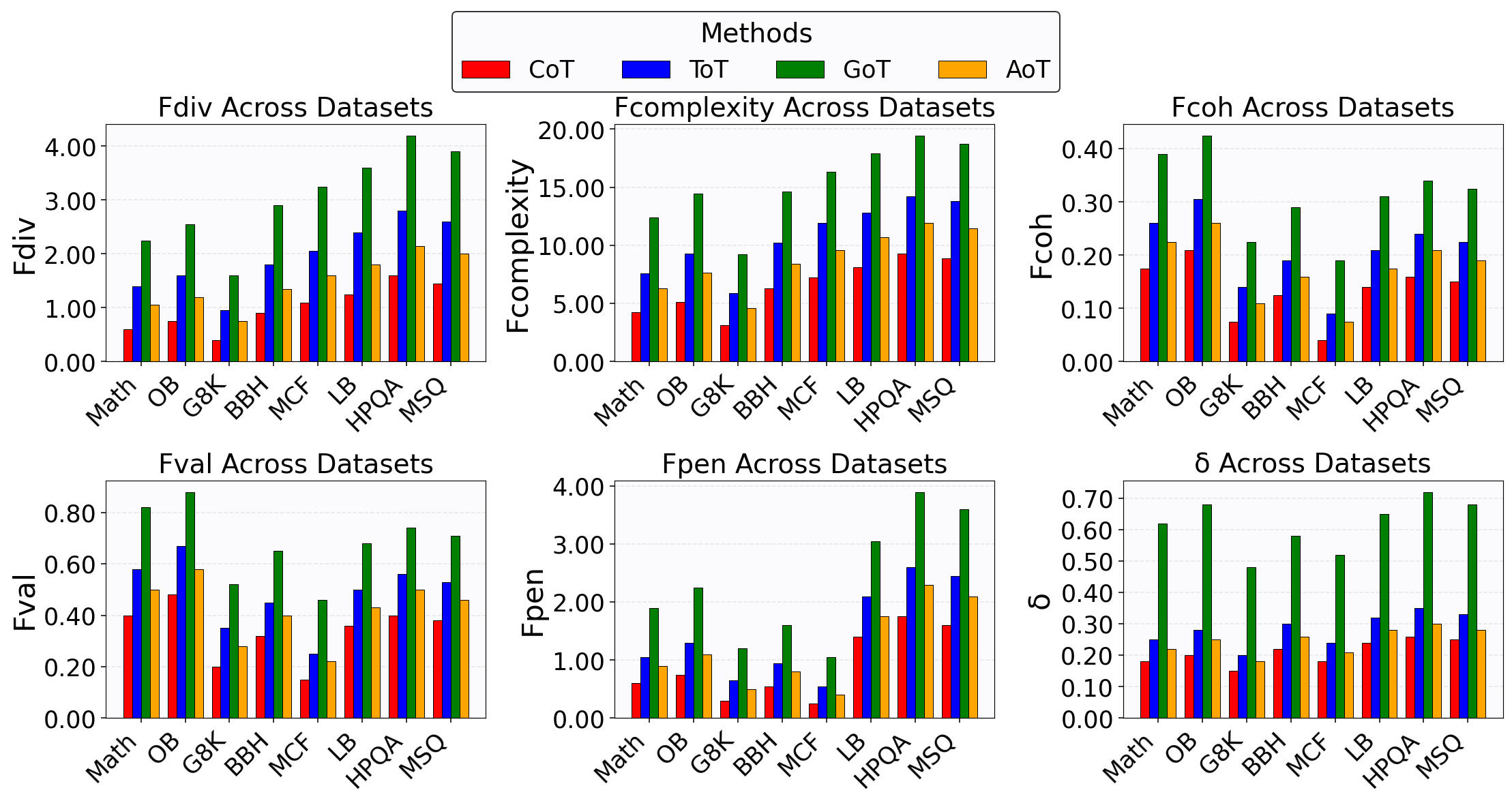}
    \caption{Topological profiles of six metrics, averaged across datasets for CoT, ToT, GoT, and AoT, computed via TDA.}
    \label{fig:line6}
\end{figure}

To characterize the structural behavior of different reasoning paradigms, we compute the six TDA metrics for all reasoning chains across datasets and then average them within each paradigm. As shown in Figure~\ref{fig:line6}, high-performing multi-step paradigms (ToT, GoT, AoT) exhibit highly consistent topological profiles, whereas CoT systematically deviates from this stable region. Specifically, CoT shows lower validation rate ($F_{val}$), loop coherence ($F_{coh}$), structural complexity ($F_{complexity}$), and penetrability ($F_{pen}$), indicating shallow reasoning depth, limited verification, and weak local consistency. Its branch diversity ($F_{div}$) and graph density ($\delta$) are also substantially smaller, reflecting an overly linear and structurally impoverished reasoning pattern. These discrepancies reveal that the gap between CoT and multi-round paradigms is not only semantic but also reflected as a consistent shift in the underlying topology. The cross-dataset convergence of multi-step paradigms suggests that effective reasoning chains cluster within a compact and quantifiable “structural health region’’ rather than varying arbitrarily. 
This observation motivates the construction of task-specific health bands: the empirical
distribution of the six TDA metrics reveals consistent structural regularities shared by strong reasoning paradigms and thus provides a natural basis for defining structurally valid regions. Paradigm-level trends outline where effective reasoning tends to lie, but these averaged profiles lack the resolution needed for task-level constraints. A statistically grounded description requires returning to individual reasoning chains, and only correct traces reflect structurally valid solution paths. Hence, the health bands should be derived from the trace-level distribution of these reliable instances.

\begin{figure}[h]
    \centering
    \includegraphics[width=1\linewidth]{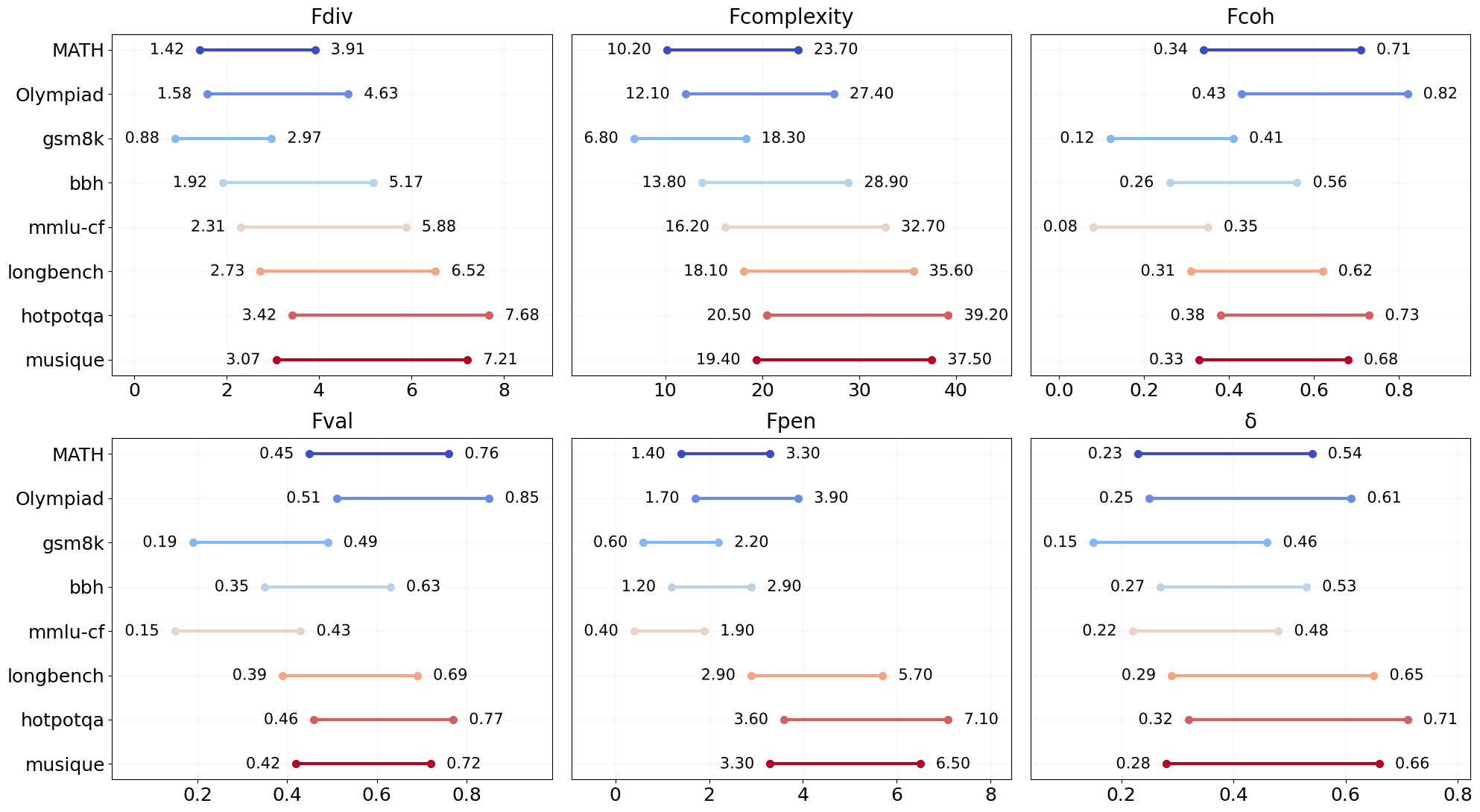}
    \caption{Task-specific Topological Health Bands derived from core topological metrics computed on correct reasoning traces.}
    \label{fig:top}
\end{figure}

Guided by this principle, we estimate the task-specific health bands directly from the
trace-level statistics. For each dataset (e.g., GSM8K, MATH, HotpotQA), we retain only the correctly solved instances from each reasoning paradigm and compute the six metrics for every trace. For each metric $F_i$, we take its 25th and 75th percentiles ($Q_1$ and $Q_3$) and define the corresponding health interval $H^{(T)}_i = [Q_1, Q_3]$. This inter-quartile region captures the middle 50\% of the distribution, yielding a robust estimate of the stable structural patterns while filtering out collapsed or highly divergent reasoning behaviors. Figure~\ref{fig:top} visualizes the empirical distributions of representative metrics across datasets. The resulting health bands characterize the structurally reasonable configurations for each task and reveal where effective reasoning processes tend to concentrate in topological space. 

All task-level bands are stored in a health base $K$ and serve as priors for the Topological Optimization Agent. During diagnosis, the agent compares a given reasoning chain’s feature vector against $H^{(T)}_i$ to compute standardized deviations $e_i$, determining whether structural repair is required and how severe the deviation is. The health bands act as safety boundaries rather than prescriptive templates, and intervention is triggered only when the reasoning topology departs significantly from the stable region. Ablation studies in Section~\ref{sec:health-threshold} further validate the effectiveness of using the $[Q_1, Q_3]$ configuration.

\begin{table}[h]
    \centering
    \caption{Spearman correlation matrix among TDA structural metrics (computed on correct reasoning traces of GSM8K and MATH combined).}
    \label{tab:tda_correlation}
    \setlength{\tabcolsep}{8pt}
    \begin{tabular}{lcccccc}
    \toprule
    Metric & $F_{\text{coh}}$ & $F_{\text{div}}$ & $F_{\text{val}}$ & $F_{\text{pen}}$ & $\delta$ & $F_{\text{complexity}}$ \\
    \midrule
    $F_{\text{coh}}$        & 1.00 & 0.18 & 0.22 & 0.11 & 0.27 & 0.19 \\
    $F_{\text{div}}$        & 0.18 & 1.00 & 0.31 & 0.09 & 0.33 & 0.42 \\
    $F_{\text{val}}$        & 0.22 & 0.31 & 1.00 & 0.28 & 0.21 & 0.36 \\
    $F_{\text{pen}}$        & 0.11 & 0.09 & 0.28 & 1.00 & 0.25 & 0.29 \\
    $\delta$                & 0.27 & 0.33 & 0.21 & 0.25 & 1.00 & 0.31 \\
    $F_{\text{complexity}}$ & 0.19 & 0.42 & 0.36 & 0.29 & 0.31 & 1.00 \\
    \bottomrule
    \end{tabular}
\end{table}

\begin{table*}[h]
  \centering
  \small
  \caption{%
  Accuracy of prompt-based optimization methods across multiple datasets using three LLMs.
  }
  \label{tab:prompt_methods}
  \setlength{\tabcolsep}{4pt}
  \renewcommand{\arraystretch}{0.9}
  \begin{adjustbox}{max width=\textwidth}
    \begin{tabular}{lccccccccc}
      \toprule
      \textbf{Method} &
      \textbf{MATH (\%)} &
      \textbf{OlympiadBench (\%)} &
      \textbf{GSM8K (\%)} &
      \textbf{BBH (\%)} &
      \textbf{MMLU-CF (\%)} &
      \textbf{LongBench (\%)} &
      \textbf{HotpotQA (\%)} &
      \textbf{MuSiQue (\%)} &
      \textbf{Avg (\%)} \\
      \midrule

      \rowcolor{gray!10}\multicolumn{10}{c}{\textbf{GPT-4o-mini}} \\
      HoT & 80.8 & 11.7 & 91.6 & 80.1 & 70.0 & 58.4 & 68.8 & 36.9 & 62.3 \\
      Instruction Induction & 81.3 & 11.9 & 92.9 & 80.5 & 71.0 & 58.8 & 69.7 & 36.5 & 62.8 \\
      Role / Persona Prompting & 80.5 & 11.3 & 92.4 & 79.9 & 70.8 & 58.6 & 69.2 & 35.9 & 62.3 \\
      Prompt Canvas & 80.1 & 11.0 & 92.1 & 79.6 & 70.6 & 58.5 & 68.7 & 35.5 & 62.0 \\
      Analogical Prompting & 65.4 & 6.5 & 87.2 & 72.5 & 65.8 & 52.9 & 64.7 & 32.8 & 56.0 \\
      TDA-RC (ours) & \textbf{82.7} & \textbf{12.6} & \textbf{94.2} & \textbf{82.5} & \textbf{72.1} & \textbf{59.3} & \textbf{70.6} & \textbf{37.8} & \textbf{63.9} \\

      \midrule
      \rowcolor{gray!10}\multicolumn{10}{c}{\textbf{Qwen-Turbo}} \\
      HoT & 80.7 & 11.4 & 91.4 & 80.0 & 69.9 & 58.2 & 69.0 & 36.6 & 62.1 \\
      Instruction Induction & 80.9 & 11.3 & 92.5 & 80.1 & 70.6 & 58.3 & 69.1 & 35.8 & 62.3 \\
      Role / Persona Prompting & 80.1 & 10.7 & 92.0 & 79.5 & 70.4 & 58.1 & 68.6 & 35.2 & 61.8 \\
      Prompt Canvas & 79.7 & 10.4 & 91.7 & 79.2 & 70.2 & 58.0 & 68.1 & 34.8 & 61.5 \\
      Analogical Prompting & 65.2 & 6.2 & 87.0 & 72.2 & 65.2 & 52.7 & 64.5 & 32.6 & 55.7 \\
      TDA-RC (ours) & \textbf{82.4} & \textbf{12.3} & \textbf{93.7} & \textbf{82.2} & \textbf{71.5} & \textbf{59.0} & \textbf{70.0} & \textbf{37.0} & \textbf{63.5} \\

      \midrule
      \rowcolor{gray!10}\multicolumn{10}{c}{\textbf{DeepSeek-V3}} \\
      HoT & 81.0 & 12.0 & 92.0 & 80.4 & 70.4 & 58.6 & 69.6 & 37.2 & 62.6 \\
      Instruction Induction & 81.6 & 12.2 & 93.4 & 80.8 & 71.4 & 59.0 & 70.0 & 36.7 & 63.1 \\
      Role / Persona Prompting & 80.7 & 11.5 & 92.8 & 80.1 & 71.1 & 58.7 & 69.4 & 36.0 & 62.5 \\
      Prompt Canvas & 80.2 & 11.1 & 92.4 & 79.7 & 70.8 & 58.5 & 68.8 & 35.5 & 62.1 \\
      Analogical Prompting & 65.6 & 6.7 & 87.6 & 72.8 & 66.1 & 53.4 & 64.9 & 33.1 & 56.3 \\
      TDA-RC (ours) & \textbf{82.9} & \textbf{12.8} & \textbf{94.4} & \textbf{82.9} & \textbf{73.2} & \textbf{59.5} & \textbf{70.8} & \textbf{37.9} & \textbf{64.2} \\

      \bottomrule
    \end{tabular}
  \end{adjustbox}
\end{table*}

\begin{table*}[h]
  \centering
  \small
  \caption{%
  Accuracy of chain-based reasoning enhancement methods across multiple datasets using three LLMs.
  }
  \label{tab:chain_methods}
  \setlength{\tabcolsep}{4pt}
  \renewcommand{\arraystretch}{0.9}
  \begin{adjustbox}{max width=\textwidth}
    \begin{tabular}{lccccccccc}
      \toprule
      \textbf{Method} &
      \textbf{MATH (\%)} &
      \textbf{OlympiadBench (\%)} &
      \textbf{GSM8K (\%)} &
      \textbf{BBH (\%)} &
      \textbf{MMLU-CF (\%)} &
      \textbf{LongBench (\%)} &
      \textbf{HotpotQA (\%)} &
      \textbf{MuSiQue (\%)} &
      \textbf{Avg (\%)} \\
      \midrule

      \rowcolor{gray!10}\multicolumn{10}{c}{\textbf{GPT-4o-mini}} \\
      CoT & 78.3 & 9.3 & 90.9 & 78.3 & 69.6 & 57.6 & 67.2 & 34.1 & 60.7 \\
      CoT-SC (n=5) & 81.8 & 10.2 & 92.0 & 83.4 & 71.1 & 58.6 & 66.2 & 33.8 & 62.1 \\
      Self-Refine & 78.7 & 9.4 & 91.7 & 80.0 & 69.7 & 58.2 & 68.3 & 35.1 & 61.4 \\
      AFlow & 83.0 & 12.4 & 93.5 & 76.0 & 69.5 & 61.0 & 73.5 & 38.1 & 63.4 \\
      ToT & 82.3 & 11.4 & 94.9 & 84.1 & 71.6 & 62.8 & 76.8 & \textbf{39.1} & 65.4 \\
      GoT & 83.0 & \textbf{13.1} & 94.5 & 85.9 & 70.2 & 63.1 & 74.2 & 36.5 & 65.1 \\
      FoT (n=8) & 82.5 & 12.5 & 94.0 & 82.4 & 70.6 & 59.1 & 66.7 & 35.8 & 63.0 \\
      AoT & \textbf{83.6} & 12.1 & \textbf{95.0} & \textbf{86.0} & 70.9 & \textbf{68.5} & \textbf{80.6} & 38.4 & \textbf{66.9} \\
      TDA-RC (ours) & 82.7 & 12.6 & 94.2 & 82.5 & \textbf{72.1} & 59.3 & 70.6 & 37.8 & 63.9 \\      

      \midrule
      \rowcolor{gray!10}\multicolumn{10}{c}{\textbf{Qwen-Turbo}} \\
      CoT & 78.1 & 8.9 & 90.7 & 78.1 & 69.4 & 57.3 & 66.8 & 33.6 & 60.0 \\
      CoT-SC (n=5) & 81.4 & 9.9 & 91.5 & 83.2 & 70.8 & 58.4 & 65.9 & 33.5 & 61.8 \\
      Self-Refine & 78.5 & 9.4 & 91.4 & 79.8 & 69.5 & 58.0 & 68.2 & 35.0 & 61.2 \\
      AFlow & 82.4 & 12.1 & 93.1 & 75.7 & 69.3 & 60.4 & 73.2 & 37.8 & 63.0 \\
      ToT & 81.9 & 11.3 & 94.2 & 83.7 & 71.3 & 62.4 & 76.4 & 38.4 & 65.0 \\
      GoT & 82.7 & \textbf{13.0} & 93.8 & 84.9 & 70.1 & 62.8 & 74.0 & 36.4 & 64.7 \\
      FoT (n=8) & 82.2 & 12.3 & 93.9 & 82.3 & 70.4 & 59.0 & 66.4 & 35.8 & 62.8 \\
      AoT & \textbf{83.5} & 12.6 & \textbf{94.7} & \textbf{85.4} & 70.5 & \textbf{68.1} & \textbf{80.0} & \textbf{39.2} & \textbf{66.8} \\
      TDA-RC (ours) & 82.4 & 12.3 & 93.7 & 82.2 & \textbf{71.5} & 59.0 & 70.0 & 37.0 & 63.5 \\

      \midrule
      \rowcolor{gray!10}\multicolumn{10}{c}{\textbf{DeepSeek-V3}} \\
      CoT & 78.5 & 9.5 & 91.3 & 78.5 & 69.9 & 57.7 & 67.4 & 34.2 & 60.9 \\
      CoT-SC & 82.0 & 10.4 & 92.1 & 83.6 & 71.5 & 58.9 & 66.6 & 34.0 & 62.4 \\
      Self-Refine & 78.9 & 9.5 & 91.9 & 80.4 & 70.1 & 58.4 & 69.1 & 35.1 & 61.7 \\
      AFlow & 83.4 & 12.5 & 93.6 & 76.4 & 69.8 & 61.4 & 74.0 & 38.2 & 63.7 \\
      ToT & 82.5 & 11.6 & 95.0 & 84.4 & 72.0 & 63.2 & 76.9 & 39.4 & 65.6 \\
      GoT & 83.2 & \textbf{13.7} & 94.5 & \textbf{86.2} & 70.3 & 63.4 & 74.2 & 36.7 & 65.3 \\
      FoT (n=8) & 82.7 & 12.6 & 94.2 & 82.6 & 70.5 & 59.3 & 66.8 & 36.2 & 63.1 \\
      AoT & \textbf{84.0} & 13.1 & \textbf{95.1} & 86.1 & 70.8 & \textbf{68.7} & \textbf{80.6} & \textbf{39.6} & \textbf{67.3} \\
      TDA-RC (ours) & 82.9 & 12.8 & 94.4 & 82.9 & \textbf{73.2} & 59.5 & 70.8 & 37.9 & 64.2 \\
      \bottomrule
    \end{tabular}
  \end{adjustbox}
\end{table*}

\subsubsection{Supporting Evidence}
To assess whether the six proposed topological indicators exhibit substantial redundancy, we compute the Spearman correlation matrix on correct reasoning traces from GSM8K and MATH (Table~\ref{tab:tda_correlation}). The results show that the overall correlations remain in the low to moderate range, with most coefficients falling between $0.1$ and $0.4$, and no pair of indicators displaying a strong monotonic relationship. These findings indicate that the indicators indeed capture different structural facets of reasoning graphs. For example, $F_{\text{coh}}$ (loop coherence) and $F_{\text{pen}}$ (penetrability) have extremely low correlation, suggesting that the presence of stable verification loops and the overall hierarchical depth represent independent structural properties. Likewise, $F_{\text{div}}$ (branch diversity) and $F_{\text{complexity}}$ (structural complexity) only show moderate correlation, implying that branch distribution and global graph complexity, although related, are not interchangeable.

Overall, the analysis demonstrates that the six TDA-based indicators form a non-redundant and complementary set of structural features. Each indicator highlights a distinct type of reasoning imbalance, structural weakness, or organizational pattern, thereby providing a stable and comprehensive set of signals for topological diagnosis and repair.

\subsection{Performance Evaluation}

\begin{table*}[t]
  \centering
  \small
  \caption{%
    CoT baseline token consumption (tokens) and relative cost (normalized to CoT = 1.0) of different reasoning methods across datasets, evaluated using GPT-4o-mini. All error values are standard deviations from five independent experiments.
  }
  \label{tab:cot_cost}
  \begin{adjustbox}{max width=\textwidth}
    \begin{tabular}{lcccccccc}
      \toprule
      \textbf{Dataset} &
      \textbf{MATH} &
      \textbf{OlympiadBench (Math)} &
      \textbf{GSM8K} &
      \textbf{BBH} &
      \textbf{MMLU-CF} &
      \textbf{LongBench} &
      \textbf{HotpotQA} &
      \textbf{MuSiQue} \\
      \midrule

      \rowcolor{gray!10}
      \multicolumn{9}{c}{\textbf{CoT Baseline Token Consumption (tokens)}} \\
      \textbf{Tokens} &
      2250 ± 47 & 4662 ± 68 & 264 ± 16 & 665 ± 26 & 1200 ± 35 & 78 ± 9 & 1259 ± 35 & 467 ± 22 \\

      \rowcolor{gray!10}
      \multicolumn{9}{c}{\textbf{Relative Cost (Normalized to CoT = 1.0)}} \\
      CoT & 1.00 & 1.00 & 1.00 & 1.00 & 1.00 & 1.00 & 1.00 & 1.00 \\
      CoT-SC (n=5) & 5.14 ± 0.05 & 5.07 ± 0.03 & 4.93 ± 0.21 & 5.21 ± 0.13 & 5.03 ± 0.10 & 4.97 ± 0.57 & 5.12 ± 0.10 & 5.26 ± 0.18 \\
      Self-Refine & 2.63 ± 0.04 & 2.82 ± 0.03 & 2.37 ± 0.11 & 2.28 ± 0.06 & 1.83 ± 0.05 & 2.46 ± 0.27 & 2.68 ± 0.07 & 2.91 ± 0.12 \\
      AFlow & 4.64 ± 0.07 & 5.08 ± 0.05 & 3.74 ± 0.18 & 4.16 ± 0.10 & 3.18 ± 0.08 & 4.14 ± 0.49 & 4.59 ± 0.12 & 5.13 ± 0.21 \\
      ToT & 6.19 ± 0.14 & 7.03 ± 0.11 & 4.69 ± 0.32 & 5.61 ± 0.24 & 4.12 ± 0.13 & 6.08 ± 0.74 & 6.63 ± 0.20 & 7.61 ± 0.37 \\
      GoT & 6.58 ± 0.15 & 7.46 ± 0.12 & 4.97 ± 0.33 & 6.08 ± 0.25 & 4.32 ± 0.14 & 6.49 ± 0.79 & 6.87 ± 0.21 & 7.83 ± 0.39 \\
      FoT (n=8) & 8.13 ± 0.18 & 8.09 ± 0.13 & 8.02 ± 0.52 & 8.18 ± 0.36 & 7.96 ± 0.23 & 8.06 ± 1.03 & 8.04 ± 0.36 & 8.22 ± 0.63 \\
      AoT & 3.63 ± 0.05 & 3.86 ± 0.03 & 3.29 ± 0.14 & 3.33 ± 0.07 & 2.89 ± 0.06 & 3.37 ± 0.33 & 3.68 ± 0.09 & 3.82 ± 0.16 \\
      HoT & 1.29 ± 0.02 & 1.36 ± 0.02 & 1.17 ± 0.07 & 1.23 ± 0.04 & 1.12 ± 0.03 & 1.19 ± 0.14 & 1.26 ± 0.03 & 1.31 ± 0.06 \\
      Instruction Induction & 1.19 ± 0.02 & 1.23 ± 0.02 & 1.12 ± 0.07 & 1.15 ± 0.04 & 1.08 ± 0.03 & 1.13 ± 0.13 & 1.19 ± 0.03 & 1.22 ± 0.05 \\
      Role / Persona Prompting & 1.17 ± 0.02 & 1.21 ± 0.02 & 1.10 ± 0.06 & 1.12 ± 0.03 & 1.07 ± 0.03 & 1.12 ± 0.13 & 1.16 ± 0.03 & 1.19 ± 0.05 \\
      Prompt Canvas & 1.18 ± 0.02 & 1.21 ± 0.02 & 1.11 ± 0.06 & 1.12 ± 0.03 & 1.07 ± 0.03 & 1.15 ± 0.14 & 1.19 ± 0.03 & 1.22 ± 0.05 \\
      Analogical Prompting & 1.07 ± 0.02 & 1.10 ± 0.02 & 1.05 ± 0.06 & 1.04 ± 0.04 & 0.98 ± 0.03 & 1.01 ± 0.12 & 1.03 ± 0.03 & 1.09 ± 0.05 \\
      TDA-RC (ours) & 1.17 ± 0.02 & 1.21 ± 0.02 & 1.09 ± 0.06 & 1.10 ± 0.03 & 1.06 ± 0.03 & 1.10 ± 0.13 & 1.14 ± 0.03 & 1.16 ± 0.05 \\
      \bottomrule
    \end{tabular}
  \end{adjustbox}
\end{table*}

\begin{figure*}[h]
    \centering
    \begin{minipage}{0.32\textwidth}
        \centering
        \includegraphics[width=\textwidth]{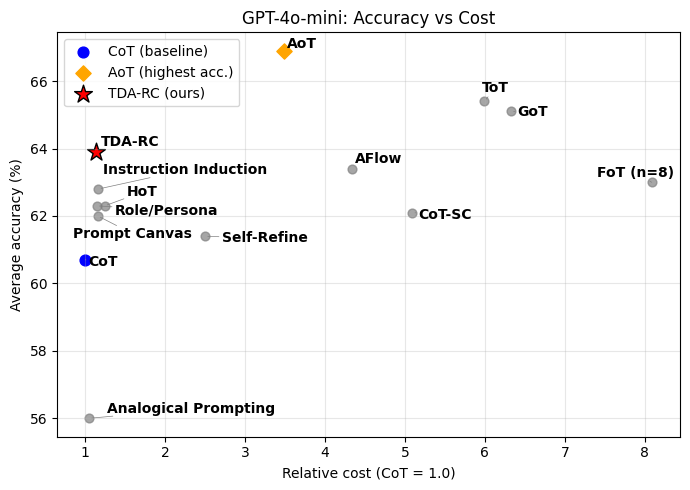}
    \end{minipage}
    \hfill
    \begin{minipage}{0.32\textwidth}
        \centering
        \includegraphics[width=\textwidth]{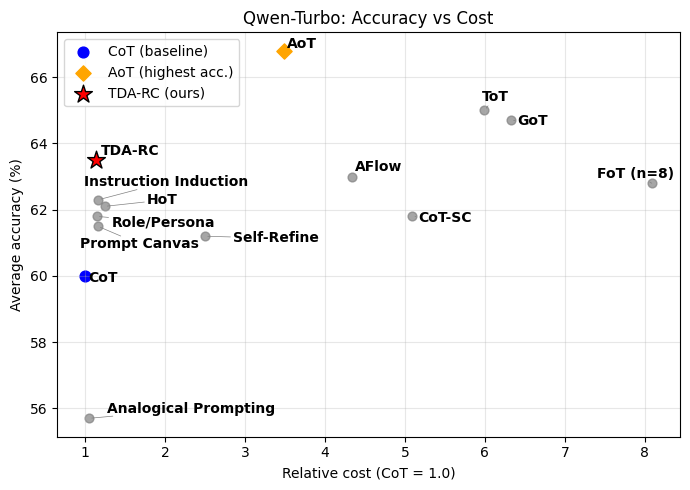}
    \end{minipage}
    \hfill
    \begin{minipage}{0.32\textwidth}
        \centering
        \includegraphics[width=\textwidth]{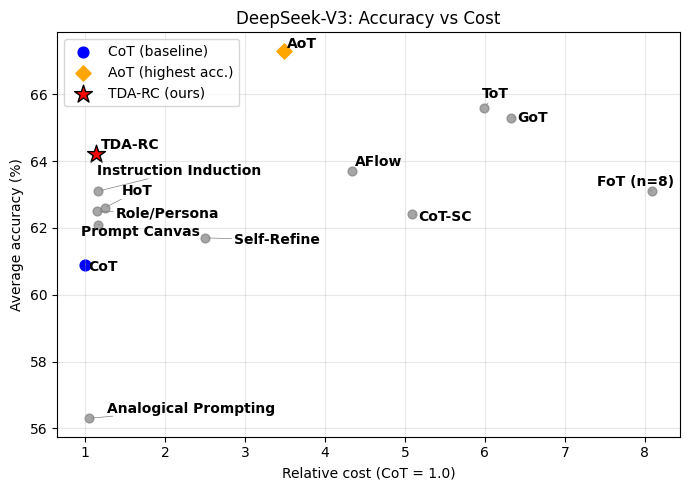}
    \end{minipage}
    \caption{Accuracy vs.\ relative cost (CoT = 1.0) for reasoning methods on GPT-4o-mini, Qwen-Turbo, and DeepSeek-V3. Each point represents a reasoning approach, with accuracy on the y-axis and generation cost on the x-axis. CoT, AoT, and TDA-RC are highlighted for reference. Methods closer to the upper-left region achieve a more favorable.}
    \label{fig:acc_cost}
\end{figure*}
In this section, we summarize the main experimental results of TDA-RC from two perspectives: reasoning accuracy and reasoning cost. Experiments are conducted on three representative LLMs (GPT-4o-mini, Qwen-Turbo, and DeepSeek-V3) to assess the generality of the proposed topological diagnostic mechanism across different architectures and task types. As shown in Tables~\ref{tab:chain_methods} and~\ref{tab:prompt_methods}, TDA-RC yields consistent accuracy improvements over standard CoT, with average gains of roughly $+3\!\sim\!+4$ points across models. In the prompt-based setting (Table~\ref{tab:prompt_methods}), TDA-RC achieves the highest accuracy in all dataset–model combinations, outperforming methods such as HoT, Instruction Induction, and Prompt Canvas. In the chain-based setting (Table~\ref{tab:chain_methods}), it remains competitive with multi-path search methods (e.g., ToT, GoT, AoT) despite operating on a single reasoning chain. This reflects the intended behavior of TDA-RC: improving the structural quality of CoT traces without relying on additional rollouts or explicit search, leading to steady but not necessarily maximal gains among chain-based approaches. Table~\ref{tab:cot_cost} further shows that TDA-RC requires only $1.06\!\sim\!1.21\times$ the token cost of CoT, while multi-round paradigms generally incur substantially higher costs ($3\!\times$–$7\!\times$). To illustrate the efficiency–performance balance, Figure~\ref{fig:acc_cost} plots average accuracy versus relative cost: multi-round methods shift toward the higher-cost region, whereas CoT remains low-cost but less accurate. TDA-RC consistently appears near the upper-left region across all models, indicating a favorable balance between accuracy and efficiency relative to both types of baselines.

Overall, TDA-RC provides consistent gains in both accuracy and efficiency. By improving the structural quality of a single-round reasoning chain through lightweight topological diagnosis and repair, it attains strong performance without the overhead of multi-round methods. This shows that topological health modeling can effectively strengthen CoT reasoning while keeping resource usage minimal.

\subsection{Topological Cross-Task Generalization}

\subsubsection{Structural Transferability}
We evaluate whether the learned health bands capture task-level structural properties rather than dataset-specific artifacts by testing whether bands learned from one dataset can be directly applied to another dataset of the same reasoning type. We study two representative pairs: GSM8K~$\rightarrow$~MATH for mathematical reasoning and HotpotQA~$\leftrightarrow$~MuSiQue for multi-hop reasoning. For each pair, we compute the health bands $H_k^{(T_s)}$ using only correct reasoning traces from the source task and apply them to the target task during diagnosis and repair. We then compare the target task’s accuracy under its own bands (self-band) and under transferred bands.

\begin{table}[h]
\centering
\small
\caption{Accuracy (\%) of TDA-RC under self-band vs.\ transferred-band settings. Small differences indicate strong cross-task transferability.}
\label{tab:cross_task_accuracy}
\begin{tabular}{lccc}
\toprule
\textbf{Target Task} & \textbf{Source Task} & \textbf{Self-band} & \textbf{Transferred} \\
\midrule
MATH       & GSM8K     & 82.7 & 82.4 \\
GSM8K      & MATH      & 94.2 & 94.1 \\
HotpotQA   & MuSiQue   & 70.6 & 70.4 \\
MuSiQue    & HotpotQA  & 37.8 & 37.6 \\
\bottomrule
\end{tabular}
\end{table}
Table~\ref{tab:cross_task_accuracy} summarizes the accuracy under self-band and transferred-band settings. Across all four transfer directions, the accuracy difference remains minimal ($\Delta\mathrm{Acc} \le 0.3$). These results indicate that health bands learned from one dataset can be largely reused by another dataset within the same reasoning category. This suggests that the bands encode stable structural regularities shared by mathematical and multi-hop reasoning tasks, rather than overfitting to dataset-specific distributions.

\subsubsection{Distributional Alignment via Wasserstein Distance}

To further assess whether structurally related tasks exhibit similar topological signatures, we compute the 1-Wasserstein distance $W_1$ between the empirical distributions of each topological metric across the same task pairs. For each metric $F_k$, let $\mathcal{D}_k^{(T_s)}$ and $\mathcal{D}_k^{(T_t)}$ denote its distributions on the source and target tasks, and compute:
\begin{equation}
W_k(T_s, T_t) = W_1\!\left(\mathcal{D}_k^{(T_s)},\, \mathcal{D}_k^{(T_t)}\right).
\end{equation}
All metrics exhibit uniformly low Wasserstein distances (0.03–0.10), as shown in Table~\ref{tab:wasserstein}, indicating that the structural distributions of correct reasoning traces are highly aligned across each task pair. This structural alignment, together with the near-identical accuracy under transferred bands, provides strong evidence that the learned health bands capture task-level structural invariants rather than dataset-specific artifacts.
\begin{table}[h]
\centering
\small
\caption{1-Wasserstein distances ($W_1$) between topological metric distributions across tasks. Lower values indicate higher structural similarity.}
\label{tab:wasserstein}
\begin{tabular}{lcccc}
\toprule
\textbf{Metric} & \textbf{G$\to$M} & \textbf{M$\to$G} & \textbf{H$\to$Q} & \textbf{Q$\to$H} \\
\midrule
$F_{\mathrm{coh}}$        & 0.07 & 0.05 & 0.04 & 0.05 \\
$F_{\mathrm{div}}$        & 0.09 & 0.06 & 0.05 & 0.06 \\
$F_{\mathrm{val}}$        & 0.05 & 0.04 & 0.03 & 0.03 \\
$F_{\mathrm{pen}}$        & 0.08 & 0.05 & 0.04 & 0.04 \\
$\delta$                  & 0.06 & 0.04 & 0.03 & 0.03 \\
$F_{\mathrm{complexity}}$ & 0.10 & 0.07 & 0.06 & 0.06 \\
\bottomrule
\end{tabular}
\end{table}

\subsection{Ablation Study}
\label{sec:health-threshold}

\subsubsection{Topological health threshold ablation} We conduct an ablation study on the selection of the topological health threshold using the GPT-4o-mini model to examine how different quantile ranges affect the model’s reasoning performance. Specifically, we compare four health interval configurations ($[Q_1,Q_3]$, $[Q_1,Q_4]$, $[Q_2,Q_3]$, and $[Q_2,Q_4]$) to evaluate their impact on diagnostic sensitivity and stability. 
As shown in Table~\ref{tab:health-threshold}, the $[Q_1,Q_3]$ configuration achieves the highest overall average accuracy (63.81\%) and exhibits consistently stable performance across tasks. This indicates that a moderately conservative quantile range provides an effective trade-off between diagnostic precision and generalization: it preserves sensitivity to meaningful topological deviations while avoiding the inclusion of noisy or overly tolerant patterns. In contrast, extending the upper bound to $Q_4$ (i.e., $[Q_1,Q_4]$) increases diagnostic tolerance but reduces stability, while narrowing the lower bound to $Q_2$ (i.e., $[Q_2,Q_3]$) limits coverage of informative structural variations. Although $[Q_2,Q_4]$ achieves a comparable average accuracy (63.78\%), it shows higher variance across datasets, suggesting reduced cross-task robustness.
\begin{table}[h]
\centering
\small
\caption{
Ablation results of different health threshold ranges in the TDA-RC framework on GPT-4o-mini. Higher values indicate better accuracy (\%).
}
\label{tab:health-threshold}
\resizebox{\linewidth}{!}{
\begin{tabular}{l
                S[table-format=2.1]
                S[table-format=2.1]
                S[table-format=2.1]
                S[table-format=2.1]}
\toprule
\textbf{Dataset} & {\textbf{[Q1,Q3]}} & {\textbf{[Q1,Q4]}} & {\textbf{[Q2,Q3]}} & {\textbf{[Q2,Q4]}} \\
\midrule
MATH                 & 82.7 & 82.2 & 82.9 & \textbf{83.0} \\
OlympiadBench (Math) & 12.6 & 12.3 & 12.8 & \textbf{12.9} \\
GSM8K                & 94.2 & 93.8 & 94.0 & \textbf{94.4} \\
BBH                  & 82.5 & 82.0 & \textbf{82.7} & 82.1 \\
MMLU-CF              & \textbf{70.8} & 70.3 & 70.6 & 70.7 \\
LongBench            & \textbf{59.3} & 58.6 & 58.8 & 58.7 \\
HotpotQA             & 70.6 & 69.9 & \textbf{70.8} & 70.5 \\
MuSiQue              & 37.8 & 37.0 & 37.5 & \textbf{38.0} \\
\midrule
\textbf{Average}     & \textbf{63.81} & 63.26 & 63.76 & 63.78 \\
\bottomrule
\end{tabular}}
\vspace{3pt}
\begin{minipage}{\linewidth}
\end{minipage}
\end{table}
Overall, these findings demonstrate that the $[Q_1,Q_3]$ interval offers the most balanced performance, effectively distinguishing healthy reasoning structures from anomalous ones while maintaining cross-task stability. Therefore, we adopt the $[Q_1, Q_3]$ configuration as the default topological health threshold in all experiments.

\subsubsection{Module-Level Ablation Study} 
To assess the independent and synergistic effects of each core component within the TDA-RC framework, we conducted a module-level ablation study. This experiment aims to clarify how specific topological signals and adaptive control mechanisms influence both reasoning accuracy and structural stability. Table~\ref{tab:ablation} summarizes the results for five configurations (A1–A5) evaluated on the GPT-4o-mini model, where each configuration systematically removes or replaces certain modules from the complete TDA-RC pipeline. Specifically, configuration \textbf{A1} represents the full framework, which integrates topological feature extraction, task preference adaptation, structural diagnosis, agent optimization, and dynamic iterative evaluation. Configuration \textbf{A2} excludes the cycle-consistency signal (–$F_{\text{coh}}$) to examine the role of loop-based structural feedback, whereas \textbf{A3} removes the penetration signal (–$F_{\text{pen}}$) to evaluate its contribution to reasoning depth and information propagation. In \textbf{A4}, the branch-diversity signal (–$F_{\text{div}}$) is omitted to analyze its effect on reasoning robustness, while \textbf{A5} disables the graph density constraint (–$\delta$) to test the importance of maintaining structural compactness. 

\begin{table}[h]
\centering
\small
\caption{Ablation study results (\%) of the proposed TDA-RC framework under different module configurations (A1–A5) on the GPT-4o-mini model. A1–A5 denote progressively modified variants removing or replacing specific modules.}
\label{tab:ablation}
\resizebox{\linewidth}{!}{
\begin{tabular}{l
                S[table-format=2.1]
                S[table-format=2.1]
                S[table-format=2.1]
                S[table-format=2.1]
                S[table-format=2.1]}
\toprule
\textbf{Dataset} & {\textbf{A1}} & {\textbf{A2}} & {\textbf{A3}} & {\textbf{A4}} & {\textbf{A5}} \\
\midrule
MATH                 & \textbf{82.7} & 81.1 & 81.2 & 81.3 & 81.4 \\
OlympiadBench (Math) & \textbf{12.6} & 11.8 & 11.9 & 12.0 & 12.1 \\
GSM8K                & \textbf{94.2} & 91.9 & 91.9 & 92.0 & 92.0 \\
BBH                  & \textbf{82.5} & 80.6 & 80.6 & 80.5 & 80.7 \\
MMLU-CF              & \textbf{70.8} & 70.4 & 70.4 & 70.4 & 70.5 \\
LongBench            & \textbf{59.3} & 58.7 & 58.6 & 58.7 & 58.7 \\
HotpotQA             & \textbf{70.6} & 70.2 & 70.1 & 70.2 & 70.3 \\
MuSiQue              & \textbf{37.8} & 37.3 & 37.2 & 37.4 & 37.5 \\
\bottomrule
\end{tabular}}
\vspace{3pt}
\begin{minipage}{\linewidth}
\footnotesize
\end{minipage}
\end{table}
The results demonstrate that the performance gains of TDA-RC arise from the synergistic interaction among multiple modules. The cycle consistency, penetration, and branch-diversity signals jointly form the structural backbone of the reasoning topology, while the density constraint further enhances global coherence and topological health. Overall, TDA-RC achieves an elegant balance between reasoning accuracy and computational efficiency through multi-dimensional topological diagnosis and lightweight structural optimization, validating the generality and robustness of the proposed framework.

\subsection{Case Studies}
\subsubsection{Incorrect-to-Correct Reasoning Cases}

To illustrate how TDA-RC identifies structural defects and guides the model toward a correct
solution, we present a representative example from the MATH (Algebra) dataset. The task is to
solve the equation $\sqrt{3x+1} = x - 1$ under a dependency-aware step-by-step prompting
setup. As shown in the initial reasoning trace, the model performs correct algebraic
manipulation but accepts both $x=0$ and $x=5$ as solutions without validating them against the
original equation. This omission results in an incorrect final answer and is reflected
topologically by the absence of coherence and validation loops.
\begin{tcolorbox}[
    float,
    floatplacement=h,
    enhanced,
    breakable,
    colback=blue!1,
    colframe=blue!50!black,
    colbacktitle=blue!12!white,
    coltitle=black,
    title={\textbf{Before TDA-RC Optimization(Incorrect $\rightarrow$ Correct)}},
    fonttitle=\bfseries\small,
    boxrule=0.4pt,
    arc=1.5pt,
    left=6pt,
    right=6pt,
    top=6pt,
    bottom=6pt,
    borderline west={3pt}{0pt}{blue!65},
    drop shadow southeast,
    sharp corners
]
\small

\textbf{Source:} MATH Dataset (Algebra)\\
\textbf{Task:} Solve the equation $\sqrt{3x+1} = x - 1$.\\
\textbf{Prompt:} Please reason step-by-step. In each step, you must explicitly cite which previous step (or condition) you used as a dependency.\\

\textbf{LLM Output:}
\begin{tcolorbox}[
    colback=white,
    colframe=black!25,
    boxrule=0.4pt,
    arc=1.5pt,
    left=6pt,
    right=6pt,
    top=6pt,
    bottom=6pt
]
\ttfamily
\textbf{Step 1:} To solve for $x$, I square both sides of the equation to eliminate the radical. \textcolor{gray}{Dependency: [Problem Statement]}\\

\textbf{Step 2:} Squaring gives $(\sqrt{3x+1})^2 = (x-1)^2$, which simplifies to $3x+1 = x^2 - 2x + 1$. \textcolor{gray}{Dependency: [Step 1]}\\

\textbf{Step 3:} I move all terms to one side to form a quadratic equation: $0 = x^2 - 5x$.
\textcolor{gray}{Dependency: [Step 2]}\\

\textbf{Step 4:} I factor the quadratic equation as $x(x-5) = 0$.
\textcolor{gray}{Dependency: [Step 3]}\\

\textbf{Step 5:} This yields two possible values for $x$: $0$ and $5$.
\textcolor{gray}{Dependency: [Step 4]}\\

\textbf{Final Answer:} \textcolor{red!70!black}{$x = 0, 5$.} \wrong

\end{tcolorbox}
\end{tcolorbox}

\begin{tcolorbox}[
    float,
    floatplacement=h,
    enhanced,
    breakable,
    colback=blue!1,
    colframe=blue!50!black,
    colbacktitle=blue!12!white,
    coltitle=black,
    title={\textbf{After TDA-RC Optimization}},
    fonttitle=\bfseries\small,
    boxrule=0.4pt,
    arc=1.5pt,
    left=6pt,
    right=6pt,
    top=6pt,
    bottom=6pt,
    borderline west={3pt}{0pt}{blue!65},
    drop shadow southeast,
    sharp corners
]
\small

\textbf{Source:} MATH Dataset (Algebra)\\
\textbf{Task:} Solve the equation $\sqrt{3x+1} = x - 1$.\\
\textbf{Prompt:} Your previous reasoning result was ``$x=0, 5$''.
Structural Diagnosis: Your reasoning path was linear and lacked topological loops ($F_{coh}=0$), which often leads to extraneous roots in radical equations.
Optimization Strategy:1. Keep your derivation steps. 2. Form a closed loop: For each candidate solution, substitute it into the original equation $\sqrt{3x+1}=x-1$. 3. Check whether LHS equals RHS and discard any candidate that fails this loop test.

\textbf{LLM Output:}
\begin{tcolorbox}[
    colback=white,
    colframe=black!25,
    boxrule=0.4pt,
    arc=1.5pt,
    left=6pt,
    right=6pt,
    top=6pt,
    bottom=6pt
]
\ttfamily
\textbf{Step 1:} Square both sides: $3x+1 = x^2 - 2x + 1$. \\
\textcolor{gray}{Dependency: [Problem Statement]}\\

\textbf{Step 2:} Rearrange to standard form: $x^2 - 5x = 0$, so $x(x-5)=0$.\\
\textcolor{gray}{Dependency: [Step 1]}\\

\textbf{Step 3:} Identify candidate roots: $x_1=0$ and $x_2=5$.\\
\textcolor{gray}{Dependency: [Step 2]}\\

\textbf{Step 4 (Validation Loop 1):} Test $x=0$ in the original equation.
LHS $= \sqrt{3(0)+1} = 1$. RHS $= 0 - 1 = -1$.\\
\textcolor{gray}{Dependency: [Step 3, Problem Statement] $\leftarrow$ (Loop Formed)}\\

\textbf{Step 5 (Decision 1):} Since $1 \neq -1$, $x=0$ is an extraneous solution. Reject it.\\
\textcolor{gray}{Dependency: [Step 4]}\\

\textbf{Step 6 (Validation Loop 2):} Test $x=5$ in the original equation.
LHS $= \sqrt{3(5)+1} = 4$. RHS $= 5 - 1 = 4$.\\
\textcolor{gray}{Dependency: [Step 3, Problem Statement] $\leftarrow$ (Loop Formed)}\\

\textbf{Step 7 (Decision 2):} Since $4 = 4$, $x=5$ is valid.\\
\textcolor{gray}{Dependency: [Step 6]}\\

\textbf{Final Answer:} \textcolor{green!40!black}{The only valid solution is $x=5$.} \correct
\end{tcolorbox}
\end{tcolorbox}

TDA-RC performs structural diagnosis by comparing the reasoning trace’s TDA metrics
against the MATH-specific health bands. Table~\ref{tab:fig1} shows that several metrics fall
outside their healthy ranges, most notably $F_{\text{coh}}$ and $F_{\text{val}}$, both equal to
zero. These values indicate a purely linear chain lacking the loop structures required for
post-squaring verification. The system therefore prescribes a targeted repair strategy: preserve
the algebraic derivation but introduce explicit substitution-based validation loops for each
candidate root.
\begin{table}[h]
\centering
\small
\caption{Topological reasoning metrics for the MATH (Algebra) task, reporting initial values, task-specific health bands, and the post-optimization improvements. Missing optimized values indicate metrics not directly targeted during the refinement process.}
\label{tab:fig1}
\resizebox{\linewidth}{!}{%
\begin{tabular}{cccc}
\hline
\textbf{Topological Metric} &
\textbf{Initial Value} &
\textbf{Health Band} &
\textbf{Optimized} \\
\hline
$F_{\text{coh}}$        & 0.00       & $[0.34,\, 0.71]$                 & 0.65 \\
$F_{\text{val}}$        & 0.00       & $[0.45,\, 0.76]$                 & 0.56 \\
$F_{\text{div}}$        & 1.00       & $[1.42,\, 3.91]$                 & 2.31  \\
$F_{\text{pen}}$        & 4.20       & $[1.40,\, 3.30]$                 & 3.10 \\
$\delta$                & 0.22       & $[0.23,\, 0.54]$                 & 0.26  \\
$F_{\text{complexity}}$ & 14.50       & $[10.20,\, 23.7]$                 & 22.10 \\
\hline
\end{tabular}
}
\end{table}

After applying this optimization, the revised reasoning trace validates both $x=0$ and $x=5$
by substituting them back into the original equation, correctly rejecting the extraneous root.
The optimized metrics indicate targeted structural improvements: loop coherence and validation
both move from $0$ into the middle of their MATH health bands, and all six indicators now fall
within their task-specific healthy intervals. This reflects the introduction of explicit
verification loops and a richer local connectivity pattern, bringing the overall topology much
closer to the profile of successful traces for this task. This example highlights that TDA-RC
not only corrects the final calculation, but also reshapes the underlying reasoning process to
better approximate task-appropriate topological patterns.

\subsubsection{Correct-to-Correct (Structural Improvement) Cases}
\begin{tcolorbox}[
    float,
    floatplacement=h,
    enhanced,
    breakable,
    colback=blue!1,
    colframe=blue!50!black,
    colbacktitle=blue!12!white,
    coltitle=black,
    title={\textbf{Before TDA-RC Optimization (Correct $\rightarrow$ Correct)}},
    fonttitle=\bfseries\small,
    boxrule=0.75pt,
    arc=2pt,
    left=8pt,
    right=8pt,
    top=6pt,
    bottom=6pt,
    borderline west={3pt}{0pt}{blue!65},
    drop shadow southeast,
    sharp corners
]
\small

\textbf{Source:} MMLU-CF (Physics / General Science)\\
\textbf{Question:} Why does ice float on liquid water?\\
\textbf{Prompt:} Please reason step-by-step. In each step, you must explicitly cite which previous step (or condition) you used as a dependency.\\

\textbf{LLM Output (Original):}
\begin{tcolorbox}[
    colback=white,
    colframe=black!25,
    boxrule=0.4pt,
    arc=1.5pt,
    left=6pt,
    right=6pt,
    top=6pt,
    bottom=6pt
]
\ttfamily
\textbf{Step 1:} According to Archimedes' principle, whether an object floats depends on the density difference between the object and the fluid.\\
\textcolor{gray}{Dependency: [Problem Statement]}\\

\textbf{Step 2:} Water is a special substance. Unlike most materials that shrink when freezing, water expands in volume when it turns into ice.\\
\fbox{\textcolor{gray}{Dependency: [Step 1]}}
 \warn\\

\textbf{Step 3:} Since the mass remains constant but the volume ($V$) increases, and density is $\rho = m/V$, the density of ice becomes lower than that of liquid water.\\
\textcolor{gray}{Dependency: [Step 2]}\\

\textbf{Step 4:} Comparing values: Ice is $\approx 0.92 g/cm^3$ and liquid water is $1.0 g/cm^3$. Since $0.92 < 1.0$, ice floats.\\
\fbox{\textcolor{gray}{Dependency: [Step 3]}} \warn \\

\textbf{Final Answer:} \textcolor{green!40!black}{Ice floats because it is less dense than liquid water due to anomalous expansion.} \correct
\end{tcolorbox}
\end{tcolorbox}

\begin{tcolorbox}[
    float,
    floatplacement=h,
    enhanced,
    breakable,
    colback=blue!1,
    colframe=blue!50!black,
    colbacktitle=blue!12!white,
    coltitle=black,
    title={\textbf{After TDA-RC Optimization}},
    fonttitle=\bfseries\small,
    boxrule=0.75pt,
    arc=2pt,
    left=8pt,
    right=8pt,
    top=6pt,
    bottom=6pt,
    borderline west={3pt}{0pt}{blue!65},
    drop shadow southeast,
    sharp corners
]
\small

\textbf{Source:} MMLU-CF (Physics / General Science)\\
\textbf{Question:} Why does ice float on liquid water?\\
\textbf{Prompt:} Your answer is factually correct but structurally shallow ($F_{pen}=1.2$). You only explained the macro-result (density), not the cause.
Optimization Instruction: 1. Deepen the topology: Construct a causal chain from Micro-level to Macro-level. 2. Inject Evidence Nodes: You must explicitly reason about 'Hydrogen Bonds' and the 'Hexagonal Crystal Lattice' structure. 3. Explain why the lattice structure leads to volume expansion."\\

\textbf{LLM Output (Mechanism-Based Reasoning):}
\begin{tcolorbox}[
    colback=white,
    colframe=black!25,
    boxrule=0.4pt,
    arc=1.5pt,
    left=6pt,
    right=6pt,
    top=6pt,
    bottom=6pt
]
\ttfamily
\textbf{Step 1 (Micro-Mechanism)}: Consider the molecular properties of water ($H_2O$). Water molecules are polar and attract each other via Hydrogen Bonds.\\
\textcolor{gray}{Dependency: [Problem Statement]}\\

\textbf{Step 2 (Structural Transition):} As temperature drops below $0^\circ C$, thermal motion decreases. The Hydrogen bonds lock the molecules into a rigid Hexagonal Crystal Lattice.\\
\textcolor{gray}{Dependency: [Step 1]}\\

\textbf{Step 3 (Geometric Consequence):} This lattice structure is an 'open cage' geometry, which forces molecules to stay further apart compared to the chaotic packing in liquid water.\\
\textcolor{gray}{Dependency: [Step 2]}\\

\textbf{Step 4 (Macro-Link):} Because the molecules are held further apart, the total volume of the substance increases (expands) by about 9\%.\\ 
 \textcolor{gray}{Dependency: [Step 3]}\\

\textbf{Step 5 (Density Calculation):} Since Mass is constant and Volume increases, Density ($\rho = \frac{m}{V}$) decreases. $\rho_{ice} < \rho_{water}$.\\
\textcolor{gray}{Dependency: [Step 4]}\\

\textbf{Step 6 (Final Phenomenon):} Due to lower density, Archimedes' principle dictates that ice floats.\\
\textcolor{gray}{Dependency: [Step 5]}\\

\textbf{Final Answer:} \textcolor{green!40!black}{Ice floats because Hydrogen bonds form an open lattice structure that increases volume and lowers density.} \correct
\end{tcolorbox}
\end{tcolorbox}
We also analyze cases where the model’s answer is correct but the reasoning structure deviates
from the task’s healthy topology. The following example from MMLU-CF (Physics/General
Science) demonstrates such a situation. Although the model correctly explains that ice floats
because it is less dense than liquid water, the original reasoning chain contains structural
issues, highlighted in the example by the boxed warnings. The first issue is incorrect dependency specification. As shown in the boxed marker in
Step 2, the fact that water expands upon freezing is incorrectly made dependent on Step 1
(Archimedes’ principle), even though these statements are logically unrelated. A second boxed
warning appears in Step 4, where the numerical density comparison is incorrectly attached to
Step 3 rather than introduced as an independent evidence node. These false dependencies force
an overly linear topology and suppress the branching structure characteristic of mechanistic
scientific explanations. In addition, the reasoning lacks micro-level causal grounding, jumping
directly from macroscopic facts to the final outcome.
\begin{table}[h]
\centering
\small
\caption{Topological reasoning metrics for the MMLU-CF (Physics / General Science) task, reporting
initial values, task-specific health bands, and the corresponding optimized improvements.}
\label{tab:fig2}
\resizebox{\linewidth}{!}{%
\begin{tabular}{cccc}
\hline
\textbf{Topological Metric} &
\textbf{Initial Value} &
\textbf{Health Band} &
\textbf{Optimized} \\
\hline
$F_{\text{coh}}$       & 0.80     & $[0.08,\, 0.35]$              & 0.28 \\
$F_{\text{val}}$     & 0.85     & $[0.15,\, 0.43]$    & 0.39 \\
$F_{\text{div}}$     & 1.30     & $[2.31,\, 5.88]$              & 4.10 \\
$F_{\text{pen}}$         & 1.20     & $[0.40,\, 1.90]$        & 1.40 \\
$\delta$         & 0.22     & $[0.22,\, 0.48]$              & 0.25 \\
$F_{\text{complexity}}$    & 4.80     & $[16.20,\, 32.70]$       & 22.10 \\
\hline
\end{tabular}
}
\end{table}

These structural deficiencies correspond to the irregular TDA metrics in
Table~\ref{tab:fig2}, where $F_{\text{div}}$ and $F_{\text{complexity}}$ are notably below
the Physics-specific ranges and $F_{\text{pen}}$ reflects limited depth. TDA-RC therefore
issues a concise structural enhancement instruction that introduces key mechanistic evidence
nodes—hydrogen bonds, the hexagonal lattice, and the geometric origin of volume expansion.
After optimization, the reasoning forms a clear micro-to-macro chain, and the dependency
structure becomes more appropriately branched. The associated TDA metrics move consistently
into the Physics task's healthy bands, with particularly large gains for $F_{\text{div}}$ and
$F_{\text{complexity}}$. This demonstrates that, even when the original answer is already
correct, TDA-RC enhances the underlying structure to better approximate task-typical reasoning
patterns without redundant elaboration.

\section{Conclusion}
TDA-RC offers a topology-oriented perspective on improving reasoning quality in large language models. By regulating the global structure of reasoning chains through persistent-homology-based diagnostics, the framework enhances the stability and coherence of single-pass reasoning while maintaining low computational cost. The results highlight the value of structural priors and topological constraints as complementary signals to conventional semantic supervision, suggesting a promising direction for developing more interpretable and structurally grounded reasoning systems. Looking forward, an important research avenue is the automatic discovery and alignment of topological archetypes to strengthen the zero-shot generalizability of TDA-RC. Another direction is dynamic health-band adaptation, enabling real-time adjustment of structural constraints without reliance on precomputed statistics. These efforts may further extend the flexibility and applicability of topology-aware reasoning in future LLM systems.

\section{Limitation}
Despite its generality, TDA-RC remains bounded by a key assumption: the availability of task-specific health bands ($H_k^{(T)}$) derived from historical data. This dependence raises questions about the framework’s plug-and-play capability when confronted with entirely novel tasks that lack such priors. Although we argue that downstream tasks tend to cluster around a small set of recurring topological archetypes—and that zero-shot archetype matching can partially offset this requirement—the current framework still benefits from existing statistical profiles. We view these limitations not as fundamental obstacles but as opportunities for further exploration, particularly in automated archetype discovery, cross-task interval transfer, and integrated modeling of structural and semantic quality.

\section*{Acknowledgments}
This work was supported by the National Natural Science Foundation of China (NSFC) under the General Program (Grant No. 62572104).

This article used large language models (such as ChatGPT) as an auxiliary tool in the language polishing process, but did not use them in research conception and academic content generation.

\bibliographystyle{IEEEtran}
\bibliography{reference}
\end{document}